\documentclass[]{aircc}

\usepackage{amsmath}
\usepackage{flexisym}

\usepackage{blindtext}
\usepackage{graphicx}
\usepackage{multirow}
\usepackage{color}
\usepackage[dvipsnames]{xcolor}
\usepackage[noadjust]{cite}
\usepackage[normal, justification=centering]{caption}
\usepackage{hyperref}

\usepackage{algorithm}
\usepackage{algorithmic}
\floatname{algorithm}{Procedure}

\usepackage{bibspacing}
\makeatletter
\def\normalsize{\@setfontsize{\normalsize}{9.5bp}{12.00pt}}
\normalsize
\makeatother

\definecolor{regKmeans}{RGB}{64, 64, 64}
\definecolor{grKmeans}{RGB}{150, 0, 0}
\definecolor{gr2Kmeans}{RGB}{204, 0, 0}
\definecolor{cvKmeans}{RGB}{128, 128, 128}
\definecolor{cv2Kmeans}{RGB}{150, 150, 150}

\makeatletter
\def\@seccntformatinl#1{\csname the#1dis\endcsname\hskip 1em\relax}
\makeatother

\begin{document}

\title{Clustering for Different Scales of Measurement - the Gap-Ratio Weighted K-means Algorithm}

\author{Joris~Gu\'erin,
        Olivier~Gibaru,
        St\'ephane Thiery,
        and~Eric~Nyiri}
        
\affiliation{Laboratoire des Sciences de l'Information et des Syst\`emes (CNRS UMR 7296)\\
Arts et M\'etiers ParisTech, Lille, France\\
\small{joris.guerin@ensam.eu}}

\maketitle

\begin{abstract}
    This paper describes a method for clustering data that are spread out over large regions and which dimensions are on different scales of measurement. Such an algorithm was developed to implement a robotics application consisting in sorting and storing objects in an unsupervised way. The toy dataset used to validate such application consists of Lego bricks of different shapes and colors. The uncontrolled lighting conditions together with the use of RGB color features, respectively involve data with a large spread and different levels of measurement between data dimensions. To overcome the combination of these two characteristics in the data, we have developed a new weighted K-means algorithm, called gap-ratio K-means, which consists in weighting each dimension of the feature space before running the K-means algorithm. The weight associated with a feature is proportional to the ratio of the biggest gap between two consecutive data points, and the average of all the other gaps. This method is compared with two other variants of K-means on the Lego bricks clustering problem as well as two other common classification datasets.
\end{abstract}


\section{Introduction}

    \subsection{Motivations}

    In a relatively close future, we are likely to see industrial robots performing tasks on their own. In this perspective, we have developed a smart table cleaning application in which a robot sorts and store objects judiciously among the storage areas available. This clustering application can have different uses: workspaces can be organized before the workday, unsorted goods can be sorted before packaging, .... Even in domestic robotics, such an application, dealing with real objects, can be useful to perform household chores.
   	
   	\begin{figure}
   	\centering
   	\includegraphics[width=0.6\textwidth]{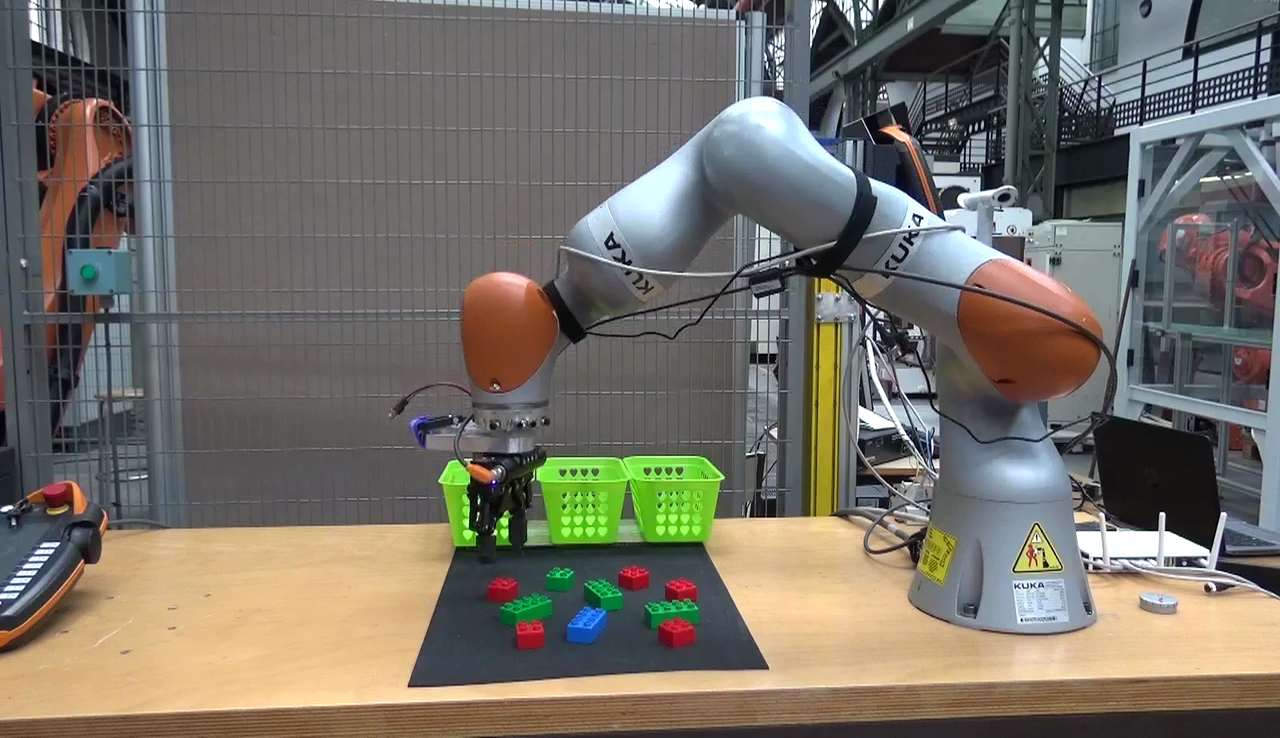}
   	\caption[KUKA LBR iiwa sorting Lego bricks]{KUKA LBR iiwa performing the Lego bricks sorting application.\\
    Video at : \url{https://www.youtube.com/watch?v=korkcYs1EHM}}
   	\label{fig:application}
   	\end{figure}
   	
    As shown in Figure \ref{fig:application}, a Kuka LBR iiwa collaborative robot equipped with a camera is presented a table cluttered with unsorted objects. Color and shape features of each object are extracted and the algorithm clusters the data in as many classes as there are storage bins. Once objects have been labelled with bin numbers, the robot physically cleans up the table. A better description of the experiment is given in the body of the article. This application was tested with Lego bricks of different colors and shapes (see section 4.2). A link to a demonstration video is given in the caption of Figure \ref{fig:application}.
     
    Because such application is meant for ordinary environments, the clustering algorithm needs to be robust to unmastered light conditions, which is synonymous with widely spread datasets. Moreover, the features chosen are on different levels of measurement \cite{level_measurement}: RGB-color features are interval type variables whereas lengths are on a ratio scale. Both these specificities of the clustering problem motivated the development of a new weighted K-means algorithm, which is the purpose of this paper.

    \subsection{Introduction to the clustering problem}
	
	The table cleaning problem described above boils down to a clustering problem \cite{pattern_recognition11, unsupervised_learning}. Clustering is a subfield of Machine Learning also called unsupervised classification. Given an unlabelled dataset, the goal of clustering is to define groups among the entities. Members in one cluster should be as similar as possible to each other and as different as possible to other clusters members. In this paper, we are only concerned with parametric clustering, where the number of clusters is a fixed parameter passed to the algorithm. However, we note that recently, non-parametric Bayesian methods for clustering have been widely studied \cite{survey_non_parametric, tuto_non_parametric, chinese_restaurant_process}.
	
	There are many possible definitions of similarity between data points. The choice of such definition, together with the choice of the metric to optimize, differentiates between the different clustering algorithms. The two surveys \cite{clustering_survey1} and \cite{clustering_survey2}, give two slightly different classifications of the various clustering algorithms.
	
	After trying several clustering algorithms on simulated Lego bricks datasets using scikit-learn \cite{sklearn}, K-means clustering

\setcounter{subsection}{0}
\setcounter{subsubsection}{0}
\section{Preliminaries}

    \subsection{K-means clustering}\label{sec2}
            
        \subsubsection{Notations} 
        
        All along this paper, we try to respect the following notations. The use of letters $i$ represents indexing on data objects whereas letter $j$ designates the features. Thus,
        \begin{itemize}
            \item $X=\{x_1, ..., x_i, ..., x_M\}$ represents the dataset to cluster, composed of $M$ data points.
            \item $F=\{f_1, ..., f_j, ..., f_N\}$ is the set of $N$ features which characterize each data object.
            \item $x_{ij}$ stands for the $j^{th}$ feature of object $x_i$.
        \end{itemize}
        A data object is represented by a vector in the feature space.
        
        Likewise, the use of letter $k$ represents the different clusters and
        \begin{itemize}
            \item $C=\{C_1, ..., C_k, ..., C_K\}$ is a set of $K$ clusters.
        \end{itemize}
        
        K-means clustering is based on the concept of centroids. Each cluster $C_k$, is represented by a cluster center, or centroid, denoted $c_k$, which is simply a point in the feature space.
        
        We also introduce $d$, the function used to measure dissimilarity between a data object and a centroid. For K-means, such dissimilarity is quantified with Euclidean distance: 
        \begin{equation}
        \label{euclidean_norm}
            d(x_i, c_k)=\sqrt{\sum_{j=1}^N (x_{ij} - c_{kj})^2}.
        \end{equation}
    
        \subsubsection{Derivation} 
        
        Given a set of cluster centers $c=\{c_1, ..., c_k, ..., c_K\}$, cluster membership is defined by
        \begin{equation}
        \label{cluster_alloc}
            x_i \in C_l \iff d(x_i, c_l) \leq d(x_i, c_k), \; \forall k \in \{1, ..., K\}.
        \end{equation}
        The goal of K-means is to find the set of cluster centers $c^*$ which minimizes the sum of dissimilarities between each data object and its closest cluster center.
        
        Introducing the binary variable $a_{ik}$, which is $1$ if $x_i$ belongs to $C_k$ and $0$ else, and the membership matrix $A = \left(a_{ik}\right)_{\substack{i \in \{1, \; \dots \; M\} \\ k \in \{1, \; \dots \; K\}}}$. K-means can be written as an optimization problem:
        \begin{equation}
        \label{optim_K_means}
            \begin{aligned}
            & \underset{A, \,\, c}{\text{Minimize}}
            & & \sum_{i=1}^M \sum_{k=1}^K a_{ik}\times d(x_i, c_k), \\
            & \text{subject to}
            & & \sum_{k=1}^K a_{ik}=1, \; \forall i \in \{1, ..., M\}, \\
            & & & a_{ik} \in \{0, \, 1\}, \; \forall i, \, \forall k.
            \end{aligned}
        \end{equation}
        
        In practice, (\ref{optim_K_means}) is optimized by solving iteratively two subproblems, one where the set $c$ is fixed and one where $A$ is fixed. The most widely used algorithm to implement K-means clustering is the Lloyd's algorithm \cite{lloyd}. It is based on the method of Alternating Optimization \cite{alternating_optimization}, also used in the Expectation-Maximization algorithm \cite{EMalgo}. The K-means optimization is composed of two main steps:
        \begin{itemize}
            \item The Expectation step (or E-step) :
            \begin{itemize}
                \item Initial situation : centroids are fixed (i.e., $c$ is fixed)
                \item Action : Each data point in $X$ is associated with a cluster following (\ref{cluster_alloc}) (i.e., $A$ is computed).
            \end{itemize}
            \item The Maximization step (or M-step) : 
            \begin{itemize}
                \item Initial situation : Each data object is associated with a given cluster (i.e., $A$ is fixed).
                \item Action : For each cluster, the centroid that minimizes the total dissimilarity within the cluster is computed (i.e., $c$ is computed).
            \end{itemize} 
        \end{itemize}
       When the norm used for dissimilarity is the $L^2$ norm, which is the case for K-means, it can be shown \cite{pattern_recognition14} that the M-step optimization is equivalent to computing the cluster mean:
       \begin{equation}
       \label{centroid_update}
           c_k = \frac{1}{\sum_{i=1}^M a_{ik}} \sum_{i=1}^M a_{ik}\times x_i.
       \end{equation}
     
    \subsubsection{Centroid initialization}
    
    In order to start iterating between the expectation and maximization steps, initial centroids need to be defined. The choice of such initial cluster centers is crucial and motivates many research, as shown in the survey paper \cite{initialization_survey}. The idea is to choose the initial centroids among the data points. In our implementation, we use K-means++ algorithm \cite{kmeans_pp} for clusters initialization (see Section 4.1).

    \subsubsection{Data normalization}
    
    In most cases, running K-means algorithm on raw data does not work well. Indeed, features with largest scales are given more importance during dissimilarity calculation and clustering results are biased. To deal with this issue, a common practice is to normalize the data before running the clustering algorithm:
    \begin{equation}
        x_{ij} \leftarrow \frac{x_{ij} - \mu_j}{\sigma_j}, \; \forall i, \, \forall j
    \label{standard_score}
    \end{equation}
    where $\mu_j$ and $\sigma_j$ represent respectively the empirical mean and variance of feature $f_j$.
    
\begin{figure}[!h]
       	\centering
       	\includegraphics[width=0.9\textwidth]{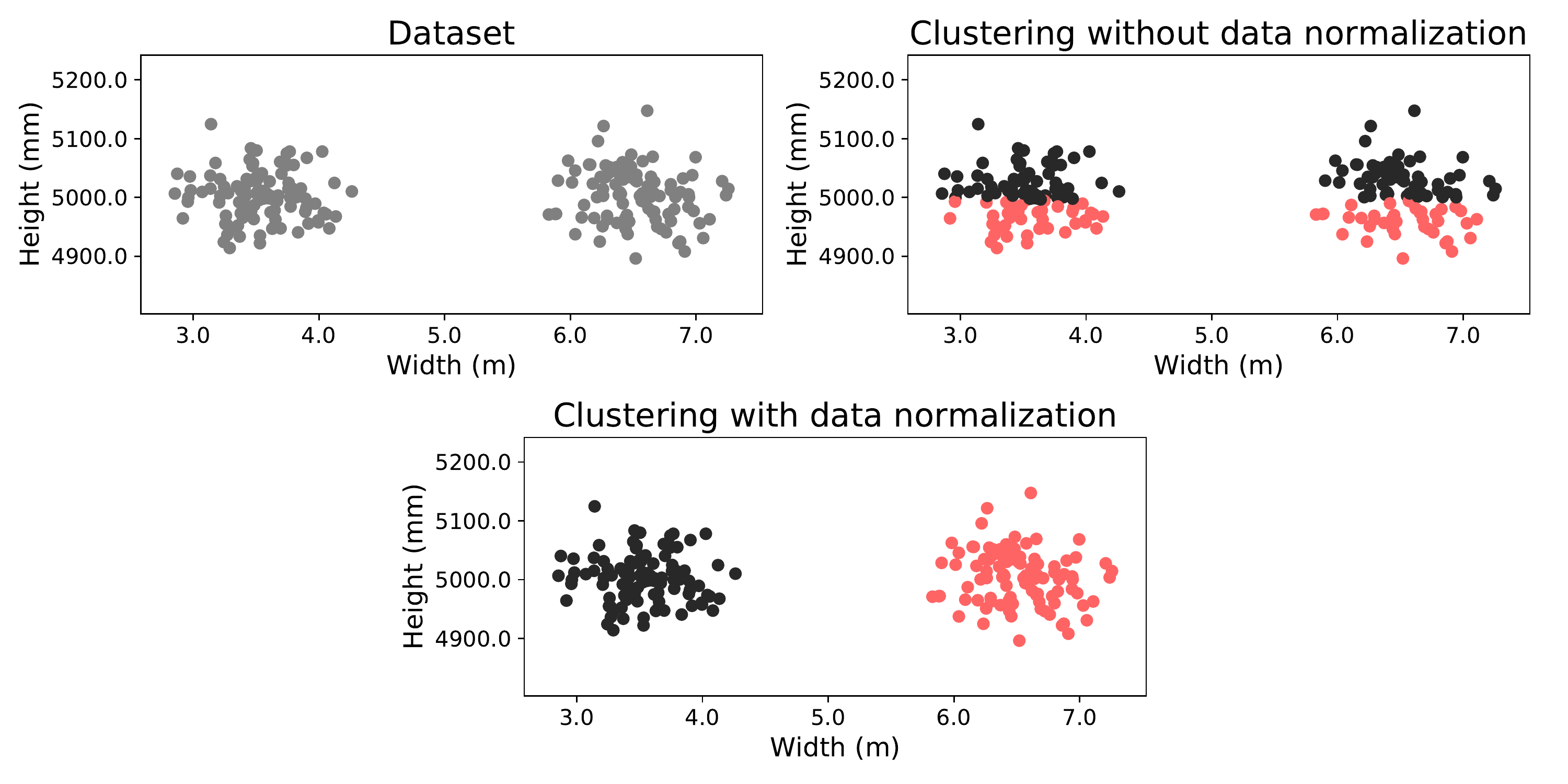}
       	\caption{Toy data set to illustrate the need for data normalization before K-means.}
       	\label{fig:example_norm}
   	\end{figure}

    The made-up, two dimensional toy dataset in Figure \ref{fig:example_norm} illustrates the interest of using data normalization as a preprocessing to K-means. The two natural clusters in Figure \ref{fig:example_norm} present similar mean and variance, but expressed in different units, which makes K-means results completely wrong without normalization.
   	
    However, reducing each feature distribution to a Gaussian of variance $1$ can involve a loss of valuable information for clustering. Weighted K-means \cite{weighted_kmeans} methods can solve such issue. The underlying idea is to capture with weights relevant information about important features. This information is reinjected in the data by multiplying each dimension with the corresponding weight after normalization. In this way, the most relevant features for clustering are enlarged and the others curtailed. In Sections 2.2.4 and 3, we propose two different weighted K-means methods: cv K-means \cite{cvKmeans} and a new method that we call gap-ratio K-means (gr K-means). These methods differ by the definition of the weights. We compare regular K-means, cv K-means and gr K-means experimentally in Section 4.
    
    \subsection{Weighted K-means}
    \setcounter{subsubsection}{0}
        \subsubsection{Issues with data normalization}
        
        As explained above, data normalization is often necessary to obtain satisfactory clustering results, but involves a loss of information that can affect the quality of the clusters found. Weighted K-means is based on the idea that information about the data can be captured before normalization and reinjected in the normalized dataset.
        
        \subsubsection{Weighted K-means}
        
        In a weighted K-means algorithm, weights are attributed to each feature, giving them different importance. Let us call $w_j$ the weight associated with feature $f_j$. Then, the norm used in the E-step of weighted K-means is:
        \begin{equation}
        \label{weighted_norm}
            d(x_i, c_k) = \sqrt{\sum_{j=1}^N w_j(x_{ij} - c_{kj})^2}
        \end{equation}
    
        The difference between weighted K-means algorithms lies in the choice of the weights.
        
        \subsubsection{Exponential weighted K-means}
        
        In this paper, we also propose an extension of weighted K-means that consists in raising the weights to the power of an integer $p$ in the norm formula:
        \begin{equation}
        \label{exp_norm}
            d(x_i, c_k) = \sqrt{\sum_{j=1}^N w_j^p(x_{ij} - c_{kj})^2}
        \end{equation}
        
        By doing so, we emphasize even more the importance of features with large weights, which makes sense if the information captured by the weights is relevant. In practice, as the weights are between 0 and 1, $p$ should not be too large to avoid considering only one feature. Influence of $p$ in the clustering results is studied in Section 4.
        
        \subsubsection{A particular example : The cv K-means algorithm}
            
        Weighted K-means based on coefficient of variation (cv K-means) \cite{cvKmeans} relies on the idea that the variance of a feature is an good indicator of its importance for clustering. Such approach makes sense, indeed, if two objects are of different nature because of a certain feature, the values of this feature come from different distributions, which increases the internal variance of the feature. In this way, the weights used for cv K-means are such that variance information is stored, so that it can be reinjected in the data after normalization.
        
        Hence, the cv weights are derived based on coefficient of variation, also called relative standard deviation. For a one dimensional dataset, it is defined by
        \begin{equation}
            cv = \frac{\sigma}{\mu}
        \end{equation}
        where $\mu$ and $\sigma$ are respectively the mean and standard deviation of the dataset (computed empirically in practice).

        Then, coefficients of variation are normalized and weights are computed such that emphasis is placed on features with highest coefficient of variation:
        \begin{equation}
            w_j = \frac{cv_j}{\sum_{j\textprime=1}^N cv_{j\textprime}}
        \label{cv_weights}
        \end{equation}
    
        cv K-means algorithm follows the same principle as regular K-means, but using norm (\ref{weighted_norm}) with weights (\ref{cv_weights}) instead of norm (\ref{euclidean_norm}).
    
        cv K-means assumes that a feature with high relative variance is more likely to involve objects being of different nature. Such approach works on several datasets but a highly noisy feature might have high variance and thwart cv K-means. However, on the original paper \cite{cvKmeans}, authors test their algorithm on three well-known classification datasets (Iris, Wine and Balance scale) from UCI repository \cite{uci_repository} and obtain better results than using regular K-means.

\setcounter{subsection}{0}
\section{Gap ratio K-means}\label{sec4}

    \subsection{Interval scale issues}

    To reason why cv weights do not fit the Lego bricks classification problem lies in the concept of levels of measurement \cite{level_measurement}. More specifically, it comes from the difference between ratio scale and interval scale.
    
    Indeed, the notion of coefficient of variation only makes sense for data on a ratio scale and does not have any meaning for data on an interval scale. On an interval scale, it is not relevant to use coefficient of variation because when the mean decreases, the variance does not change accordingly. Therefore, at equal variance, features closer to zero have higher coefficients of variation for no reason, which biases the clustering process.
    
    In the table cleaning application defined above, the features chosen are colors (RGB) and lengths. RGB-colors are given by three variables, representing the amount of red, green and blue, distributed between $0$ and $255$. They are on an interval scale and thus should not be weighted using coefficient of variation. This duality in the features measurement scales motivated the development of gap-ratio weights, which is the purpose of this section.
    
    \subsection{the gr-K-means algorithm}
    
    The idea behind gap-ratio K-means is fairly simple. When doing clustering, we want to distinguish if different feature values between two objects come from noise or from the fact that objects are of different nature. If we consider that the distribution of a certain feature differs between classes, this feature's values should be more different between objects of different classes than between objects within a class. Gap-ratio weights come from this observation, their goal is to capture this information about the features.
    
    To formulate this concept mathematically, we sort the different values $x_{ij}$ for each feature $f_j$. Hence, for every $j$, we create a new data indexing, where integers $i\{j\}$ are defined such that
    \begin{equation}
        \forall j, \, x_{i\{j\}, \, j} \leq x_{i\{j\}\textprime, \, j} \Leftrightarrow i\{j\} \leq i\{j\}\textprime.
    \end{equation}
    
    Then, we define the $i\{j\}^{th}$ gap of $j^{th}$ feature by: 
    \begin{equation}
        g_{i\{j\}, \, j} = x_{i\{j\}+1, \, j} - x_{i\{j\}, \, j}
    \end{equation}
    Obviously, if there are $M$ data objects, there are $M-1$ gaps for each feature.
    
    After computing all the gaps for feature $f_j$, we define the biggest gap $G_j$ and the average gap $\mu g_j$ as follows:
    \begin{eqnarray}
        G_j = \max_{i\{j\} \in \{1, ..., M-1\}} g_{i\{j\},j}, \nonumber\\
        \mu g_j = \frac{1}{N}\sum_{\substack{i\{j\}=1\\i\{j\} \neq I\{j\}}}^{M-1} g_{i\{j\},j},
    \end{eqnarray}
    where $I\{j\}$ is the index corresponding to $G_j$.
    
    Finally, we define the gap-ratio for feature $f_j$ by: 
    \begin{equation}
        gr_j = \frac{G_j}{\mu g_j}.
    \end{equation}
    
    In other words, for a given feature, the gap ratio is the ratio between the highest gap and the mean of all other gaps. Then, as for cv K-means, gap-ratios are used to compute scaled weights:
    \begin{equation}
    \label{weights_gr}
        w_j = \frac{gr_j}{\sum_{j\textprime=1}^N gr_{j\textprime}}.
    \end{equation}
    
    The dissimilarity measure for gr K-means is obtained by using weights (\ref{weights_gr}) in (\ref{weighted_norm}). Likewise exponential cv K-means, we call exponential gr K-means the algorithm using dissimilarity measure (\ref{exp_norm}) with weights (\ref{weights_gr}).
    
    \subsection{Intuition behind gr K-means}
    
\begin{figure}[!h]
        \centering
        \includegraphics[width = 0.9\textwidth]{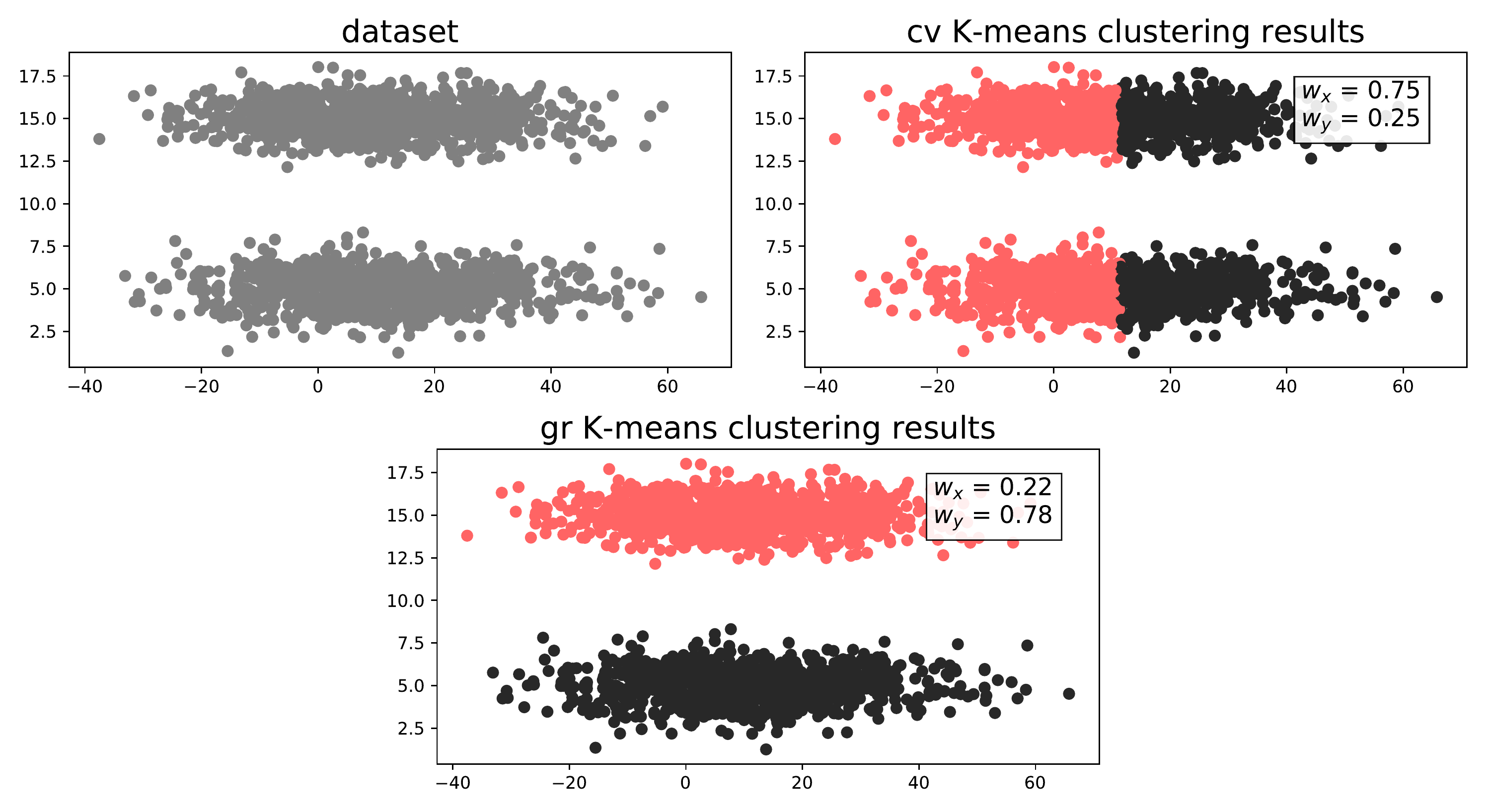}
        \caption{Comparison of cv K-means and gr K-means on a simple made up example. This is to illustrate cases where it seems more logical to deal with gaps rather than variances.}
        \label{intuition}
    \end{figure}

    Figure \ref{intuition} shows a simple two dimensional toy example where using gr weights is more appropriate than cv weights. 
    
    In this example, the coefficient of variation along the x-axis is larger than for the y-axis. Indeed, mean values for both dimensions are approximately the same (around $10$) whereas variance is higher for the x-axis. Thus, cv K-means focuses on the x-axis despite we can see it is not a good choice just by looking at the plots. The clusters found in the middle plot, together with the weights, confirm the wrong behavior of cv K-means.
    
    However, weights and groups obtained with gr K-means (bottom plot) indicate that the right information is stored in gap-ratio weights for such problem. The biggest gap along the y-axis is a lot bigger than average gaps whereas these two quantities are similar along the x-axis.

\setcounter{subsection}{0}
\setcounter{subsubsection}{0}
\section{Experimental validation}\label{sec5}
    
    In the previous sections, we have introduced the K-means clustering method. We explained why data normalization is required and why it should not work on data with relatively high spread. Then, we presented cv K-means as a solution to capture important information about the data before normalization but showed that it is not compatible with interval scale data. Finally, we derived a new weighted K-means algorithm that should fit the kind of datasets we are interested in. 
    
    In this section, we intend to validate the intuitive reasoning above. To do so, we compare the different weighted K-means algorithms (including regular K-means, with weights $w_j=1, \, \forall j$, and exponentiated weights) on different datasets. First, in Section 4.2, the Lego bricks dataset, used to demonstrate the table cleaning application, is clustered with the different methods. Then, in Section 4.3, two other famous classification datasets are used to investigate further the algorithms behaviors.
    
    \subsection{weighted K-means implementation}\label{sec5_1}
    
    In this validation section, we used the K-means implementation of scikit-learn \cite{sklearn}, an open-source library, as is. This way, our results can be checked and further improvements can be tested easily. To implement weighted K-means algorithms, we also use scikit-learn implementation but on a modified dataset. After normalization, our data are transform using the following feature map:
    
    \begin{equation}
        \Phi: x_{ij} \rightarrow \sqrt{\omega_j} x_{ij}.
    \end{equation}
    
    As the dissimilarity computation appears only in the E-step, the dataset modifications are equivalent to changing the norm. Indeed, (\ref{weighted_norm}) is the same as

    \begin{equation}
        d(x_i, c_k) = \sqrt{\sum_{j=1}^N (\sqrt{w_j}x_{ij} - \sqrt{w_j}c_{kj})^2}.
    \end{equation}
    
    By doing this, results obtained can be compared more reliably. Differences in the results is less likely to come from poor implementation as the K-means implementation used is always the same.

    \subsection{Results on the Lego classification problem}\label{sec5_2}
    
        \subsubsection{Experiment description}
    
\begin{figure}[!h]
            \centering
            \includegraphics[width = 0.35\textwidth]{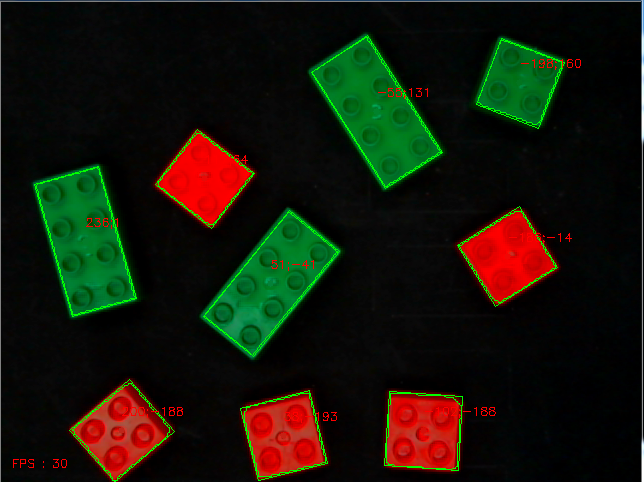}
            \includegraphics[width = 0.35\textwidth]{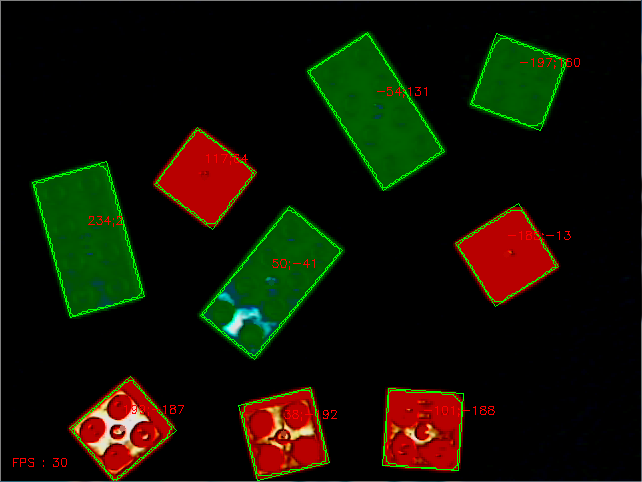}\\
            \vspace{3pt}
            \includegraphics[width = 0.35\textwidth]{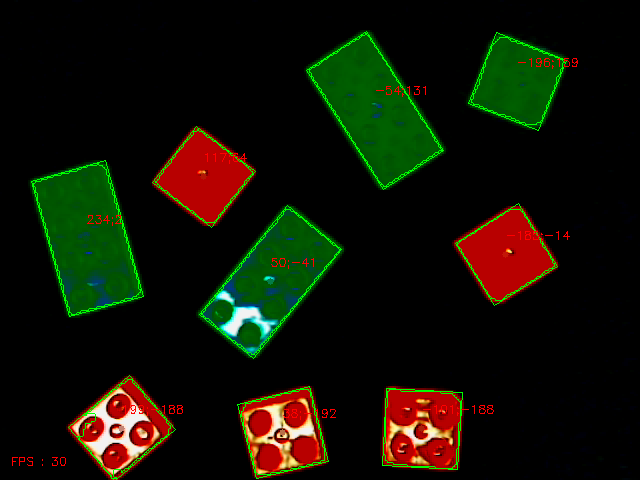}
            \includegraphics[width = 0.35\textwidth]{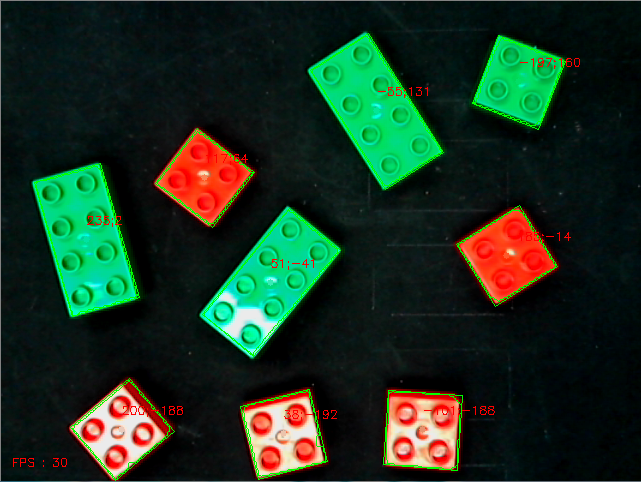}
            \caption{Data set to cluster under different light conditions.}
            \label{noisy_data}
        \end{figure}

        The first dataset used to compare different algorithms is one composed of nine Lego bricks of different sizes and colors, as shown on Figure \ref{noisy_data}. As explained in the introduction, the original goal was to develop an intelligent robot table cleaning application that can choose where to store objects using clustering. Such application is tested by sorting sets of Lego bricks because it is easy and not subjective to draw natural classes and thus validate the robot choices. Figure \ref{noisy_data} shows the kind of data sets we are dealing with, three classes can easily be found within these Lego bricks. Naturally, such set of bricks needs to be sorted within three boxes; the clustering algorithm needs to place the big green, small green and small red bricks in different bins.
        
        Furthermore, on Figure \ref{noisy_data}, we can see that among the four pictures, lighting varies a lot. Color features observed are really different between two runs of the application. The algorithm needs to be robust to poor lighting conditions and to be able to distinguish between red and green even when colors tend to be saturated (see bottom right image). 
        
        A video showing the robot application running can be found at \url{https://www.youtube.com/watch?v=korkcYs1EHM}. Three different cases are illustrated: the one of Figure \ref{noisy_data}, one with a different object (not a Lego brick), and one with four natural classes with only three boxes.
        
        The experiment goes as follows: the robot sees all the objects to cluster and extract three color features (RGB) and two length features (length and width). Colors are extracted by averaging a square of pixels around the center of the brick. The dataset gathered is then passed to all variants of weighted K-means algorithms which return the classes assigned of each object. Finally, the robot physically sorts the objects (see video). For our experimental comparison of different algorithms, the experiment has been repeated $98$ times with different arrangements of the Lego bricks presented on Figure \ref{noisy_data}, and with different lighting conditions. For each trial, if the algorithm misclassified one or more bricks, it is counted as a failure, else, it is a success.
        
        As explained in Section 4.2.2, clustering is tried with different weighted K-means algorithms but also with and without scaling. Different values of $p$ for exponentiated weighted K-means are also tried.
    
        Figure \ref{results_lego} presents results obtained on the $98$ trialsLego bricks datasets. The two left charts represent results on the original datasets and the right ones are results on the same dataset with a slight color modification. We removed $50$ to the blue component of the bricks, which corresponds to using bricks of slightly different colors, in order to test the robustness of the algorithms.
    
        \subsubsection{Results interpretation}
        
         \begin{figure}[!h]
            \centering
            \includegraphics[width = 0.45\textwidth]{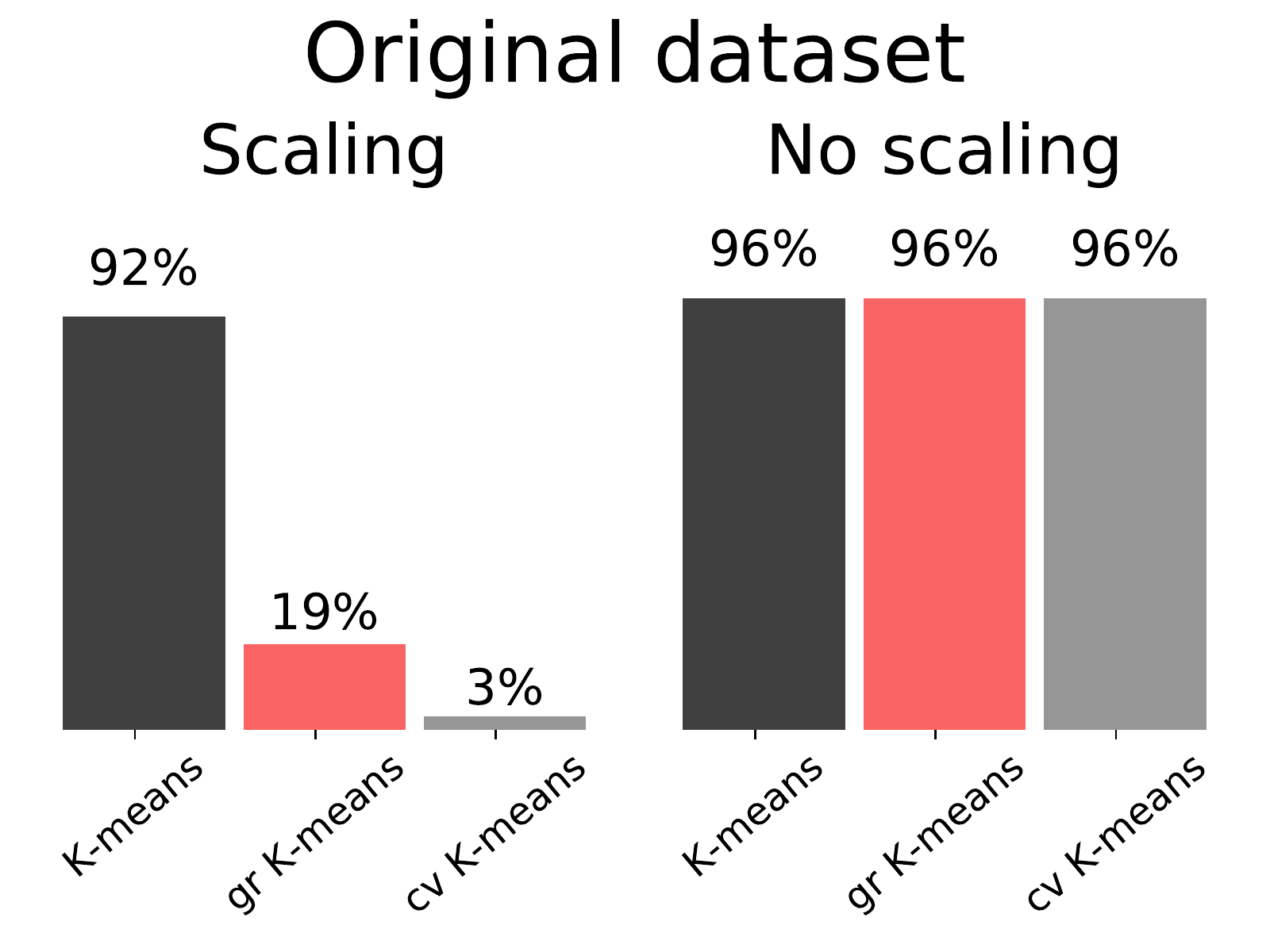}
            \hspace{3pt}
            \includegraphics[width = 0.45\textwidth]{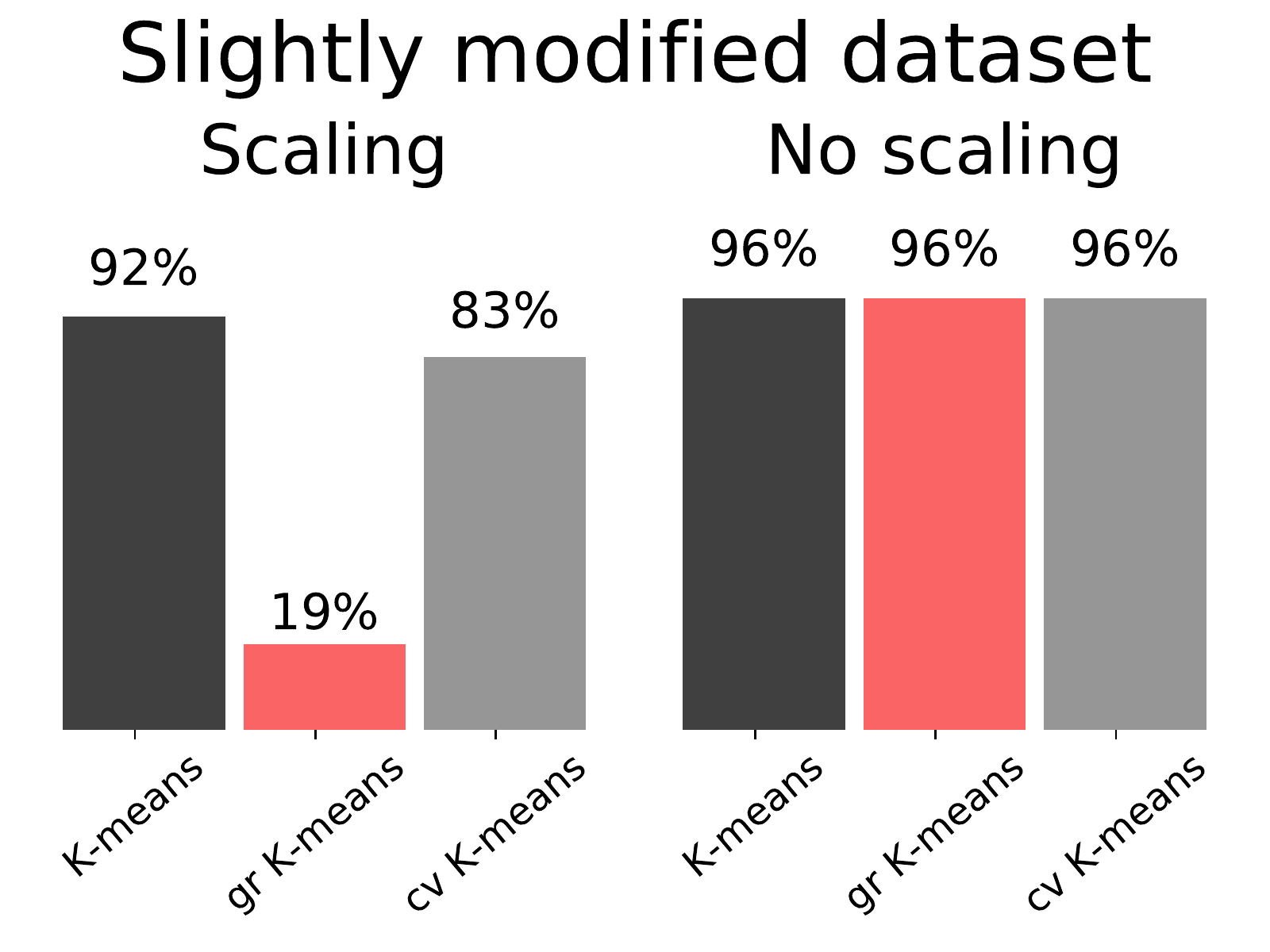}
            \caption{Percentage of experiments with at least one misclassification. The experiment was run $98$ times under different lightning conditions and with different layouts of the bricks. Error rates presented are averaged among these $98$ runs.}
            \label{results_lego}
        \end{figure}
        
        \begin{figure}[!ht]
            \centering
            \includegraphics[width = 0.95\textwidth]{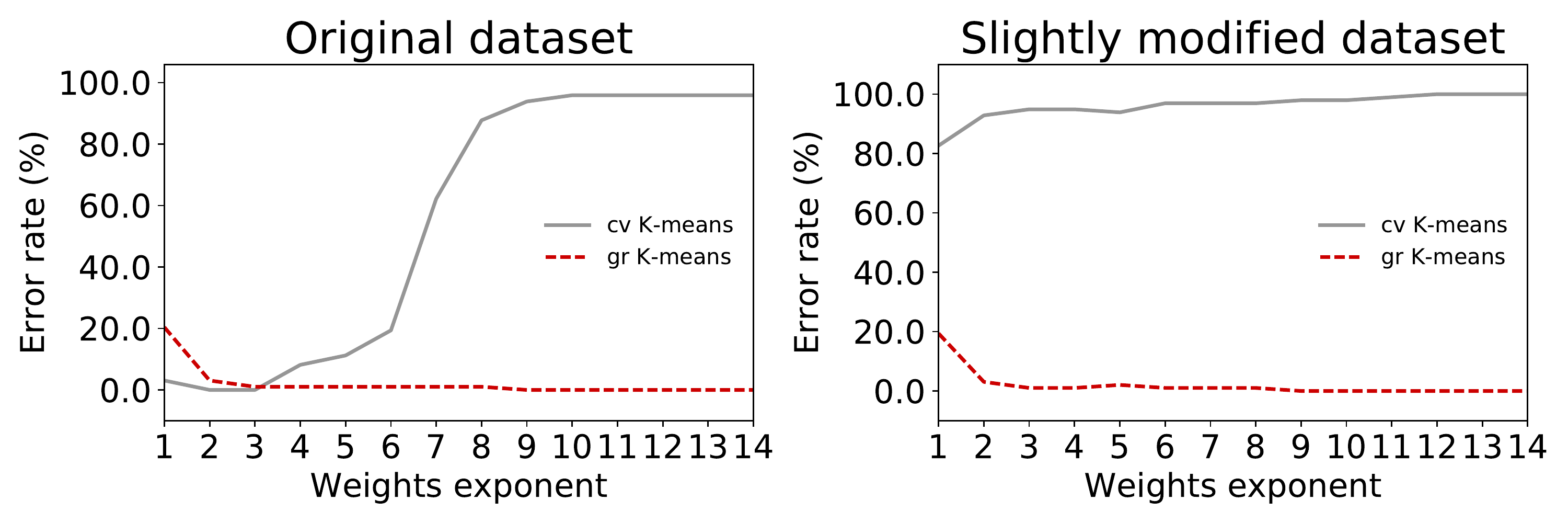}
            \caption{Exponent influence for Lego bricks clustering with exponentiated weighted K-means algorithms.}
            \label{exponent_influence}
        \end{figure}
        
        We start by analyzing the influence of scaling the dataset. As we can see on Figure \ref{results_lego}, without scaling (right column), error rates are all very large, around $95\%$. For this precise problem, K-means cannot perform good without a proper scaling of the dataset before running the clustering algorithm. Such observation makes sense as lengths ($\approx 5cm$) cannot be compared with colors ($\approx 150$) because they are totally differently scaled. K-means always put emphasis on data with the largest values (i.e., colors), which means that noise on the color has much more influence than different values of the length. Hence, for this practical example, we can assert that data scaling is required to have a decent classification.
        
        Then, let us compare the different algorithm. The first observation is that whatever preprocessing is used, regular K-means always results in very poor classification. One possible explanation for such bad behavior is the relatively high spread in the data (due to lighting conditions), which involve high variance in the features values and makes it difficult to differentiate between noise and true difference of nature. For this reason, emphasizing certain features is required and we can know compare cv and gr K-means.
        
        Looking at the left subplots of Figure \ref{results_lego}, we can see that cv K-means performs particularly well on the original dataset. However, after little modifications, it falls into very bad behavior. Such issue with cv K-means comes from the fact that coefficient of variation is not appropriate for interval scale data. In other words, cv K-means can succeed on the original dataset only because the bricks used do not present RGB components too close to zero. On the other hand, gr K-means performs reasonably well ($\approx 20\%$ error rates), and is stable to dataset modifications. 
        
        Another way to determine the relevance of the information stored in the weights is to look at different values of the exponent for exponentiated weighted K-means. Figure \ref{exponent_influence} shows such curves for both cv weights and gr weights with the original and the modified datasets. For gr weights, curves for both datasets are superimposable, gr K-means algorithm is insensitive to average values of the interval scaled features. As for cv K-means, it performs good under certain conditions (left figure) but is not robust to decreasing the mean value of one feature (bottom figure). 
        
        Figure \ref{exponent_influence} left plot shows another interesting thing. For low value of $p$, cv K-means performs better than gr K-means ($0\%$ error rate for $p=3$). Even if all we want is a certain exponent for which error is low, it is interesting to note that high exponents involve bad clustering results with cv weights. Such behavior shows that the weights are not so relevant because if they are given too much importance, clustering gets worse. On the other hand, with exponentiated gr K-means error rate tends to decrease when the exponent increases. Information carried by gr weights is good for such clustering problem and should be given more importance. Error rate falls to zero at $p=9$ and remains stable to exponent increases until relatively high values of $p$ ($>20$); the balance between important components is well respected within the weights. For this kind of datasets, characterized by large spread, mixed scales of measurement and relatively independent features, exponentiated gr K-means with relatively high exponent seems to be a good solution for clustering.   
     
    \subsection{Generalization to other data sets}\label{sec5_3}
    
    Gap-ratio weighted K-means was developed with Lego bricks classification task in mind, so it is not surprising that it performs good on such datasets. Now, we test this algorithm on other classification datasets of different nature to see how well it generalizes. Different weighted K-means methods are compared on two famous supervised learning datasets, so that we have labels to evaluate the clustering output. The two datasets chosen are the Fisher Iris dataset \cite{iris_dataset} and the Wine dataset, both taken from the UCI Machine Learning Repository \cite{uci_repository}. Table \ref{datasets_description} gives some important characteristics of both datasets.
    
    \begin{table}[!ht]
\caption{Datasets descriptions.}
\label{datasets_description}

	\centering
    \resizebox{0.6\textwidth}{!}{
    \begin{tabular}{c|c|c}
        \textbf{Dataset} & Iris & Wine \tabularnewline
        \hline
        \textbf{Number of instances} & 150 & 178 \tabularnewline
        \textbf{Number of attributes} & 4 & 13 \tabularnewline
        \textbf{Number of classes} & 3 & 3 \tabularnewline
        \textbf{Is linearly separable?} & No & Yes \tabularnewline
        \textbf{Data type} & Real & Real and Integers \tabularnewline
        \textbf{Scale of measurement} & Ratio & Ratio \tabularnewline
    \end{tabular}}
   
\end{table}
    
    \begin{table}[!ht]
\caption{Results on other data sets.} 
\label{table_result}
\centering
	\begin{minipage}{0.45\textwidth}
	\centering
    \begin{tabular}{c|c|c}
        \multicolumn{3}{c}{IRIS DATASET}\vspace{1mm}\tabularnewline
        \multicolumn{1}{c|}{} & Scaling & No scaling\tabularnewline
        \hline
         \textcolor{regKmeans}{K-means} & \textcolor{regKmeans}{17.05 \%} & \textcolor{regKmeans}{10.67 \%}\tabularnewline
         \textcolor{grKmeans}{gr K-means} & \textcolor{grKmeans}{11.73 \%} & \textcolor{grKmeans}{8.66 \%}\tabularnewline
         \textcolor{cvKmeans}{cv K-means} & \textcolor{cvKmeans}{4.12 \%} & \textcolor{cvKmeans}{5.62 \%}\tabularnewline
         \textcolor{gr2Kmeans}{gr$^2$ K-means} & \textcolor{gr2Kmeans}{4.01 \%} & \textcolor{gr2Kmeans}{5.33 \%}\tabularnewline
         \textcolor{cv2Kmeans}{cv$^2$ K-means} & \textcolor{cv2Kmeans}{4.01 \%} & \textcolor{cv2Kmeans}{4.12 \%}
    \end{tabular}
    \end{minipage}
    \begin{minipage}{0.45\textwidth}
    \centering
    \begin{tabular}{c|c|c}
        \multicolumn{3}{c}{WINE DATASET}\vspace{1mm}\tabularnewline
        \multicolumn{1}{c|}{} & Scaling & No scaling\tabularnewline
         \hline
         \textcolor{regKmeans}{K-means} & \textcolor{regKmeans}{3.36 \%} & \textcolor{regKmeans}{29.78 \%}\tabularnewline
         \textcolor{grKmeans}{gr K-means} & \textcolor{grKmeans}{5.11 \%} & \textcolor{grKmeans}{29.78 \%}\tabularnewline
         \textcolor{cvKmeans}{cv K-means} & \textcolor{cvKmeans}{7.01 \%} & \textcolor{cvKmeans}{29.78 \%}\tabularnewline
         \textcolor{gr2Kmeans}{gr$^2$ K-means} & \textcolor{gr2Kmeans}{7.87 \%} & \textcolor{gr2Kmeans}{29.78 \%}\tabularnewline
         \textcolor{cv2Kmeans}{cv$^2$ K-means} & \textcolor{cv2Kmeans}{7.30 \%} & \textcolor{cv2Kmeans}{29.78 \%}\tabularnewline
    \end{tabular}
    \end{minipage}
    
    \vspace*{3mm}
    
    \footnotesize{Error rates averaged over $1000$ runs of the algorithms from different centroid initializations. $gr^2$ and $cv^2$ denote the exponential versions of the algorithms ($p=2$).}

    \vspace*{3pt}
   
\end{table}
    
    Table \ref{table_result} summarizes clustering results for both datasets, using all previously described implementations of different algorithms. For each configuration, we ran the algorithm $1000$ times with different random initializations. The percentage reported in Table \ref{table_result} corresponds to the average error rate over the different runs.
    
    We acknowledge that data normalization decreases errors for the Wine dataset but not for the Fisher Iris dataset. We explain such results by the fact that the values of the four Iris attributes are of the same order of magnitude. Hence, normalizing involves a loss of information that is not compensated by scaling the different features.
    
    Regarding the algorithms efficiency, For the Iris dataset, both gr and cv K-means implementations are better than regular K-means. Moreover, increasing the weights exponent improves the quality of the clustering. This means that both gap-ratio and coefficient of variation weights are able to capture the important information for clustering. However, for the Wine dataset, the best option is to stick to regular K-means.
    
    Finally, we also underline that for both K-means and cv K-means, we do not find the same results than in the original paper of cv K-means (\cite{cvKmeans}). Overall we obtain lower error rates. this might come from the K-means++ initialization, as in \cite{cvKmeans}, all the points are initialized at random.
    
    In the conclusion, we propose a short recommendation section to help the reader selecting a weighted K-means algorithm given the properties of the dataset to cluster.

\setcounter{subsection}{0}
\setcounter{subsubsection}{0}
\section{Conclusion}

    \subsection{Recommendations}
    
    Preprocessing the data by normalizing the features seems to be a good idea as long as the initial dimensions present different scales. On the other hand, if features already have the same order of magnitude, it is better to leave them unchanged, unless important information is already captured, with properly chosen weights for example.
    
    As for the choice of the algorithm, from what we have observed, we suggest to stick to regular K-means when your data appears to have high correlation and clusters do not come from only a few dimensions. This is more likely to happen with high dimensional data. In contrast, on relatively low dimensional data, it seems a smart idea to go for a weighted K-means algorithm. If patterns are to be found along isolated dimensions, gap-ratio seems to be a better indicator than coefficient of variation. However, for certain cases, such as Iris dataset, we acknowledged that cv K-means produces similar results. For data on different scales of measurement, cv K-means cannot be used and gap-ratio is the right choice; especially with wide spread data.
    
    Finally, regarding weights exponentiation, we found out that for linearly separable datasets, when weighted K-means makes improvements, it is better to raise the weights to a relatively high power. Information gathered in the weights is good and should be emphasized. However, the exponent should not be too large or the algorithm ends up considering a single feature. We should remain careful to avoid loosing the multidimensionality of the problem.
    
    The algorithm developed is a new approach for clustering data that are mixed between interval and ratio measurement scales and should be considered whenever facing such case. However, as for all other clustering problems, it works only for a certain range of problems and should not be used blindly.
    
    \subsection{Future work}

    \par{\textbf{gr K-means - }} Regarding the gr K-means algorithm, we have several possibility of improvement in mind. First, combining data orthogonalization methods (such as ICA \cite{fastica}) and gap-ratio indicator seems a promising idea and it might be fruitful to search in this direction. Indeed, gr weights are computed along different dimensions of the feature space and if features have strong correlation, gaps might disappear and variance might be spread along several dimensions. For this reason, it seems appealing to try to decorrelate data using orthogonalization methods.
    
    It could also be interesting to consider not only the largest gap along one dimension but also the next ones, according to the number of different classes desired. Indeed if within three classes the two separations come from the same features, even more importance should be given to this set of features. Some modifications of the equations in Section 3 should enable to try such approach.
    
    \par{\textbf{Automatic feature extraction - }} Regarding the table cleaning application which motivated this research, developing gr K-means enabled us to get the robot sorting judiciously Lego bricks, as well as other objects (see video). However, clustering is based on carefully selected features which are only valid for a range of objects. As a future research direction, we consider trying to develop the same application with automatic feature extraction, using transfer learning from a deep convolutional network trained on a large set of images \cite{transfer_learning}.

\renewcommand{\thesection}{\hspace*{0em}}
\section{Acknowledgements}
The authors would like to thank Dr. Harsh Gazula and Pr. Hamed Sari-Sarraf for their constructive reviews of this paper.

\bibliographystyle{IEEEtran}
\bibliography{}

\end{document}